\documentclass[runningheads]{llncs}
\usepackage{graphicx}
\usepackage{amsmath,amssymb} 
\usepackage{color}

\usepackage{multirow}
\usepackage[width=122mm,left=12mm,paperwidth=146mm,height=193mm,top=12mm,paperheight=217mm]{geometry}
\usepackage{algorithm}

\newcommand\blfootnote[1]{%
	\begingroup
	\renewcommand\thefootnote{}\footnote{#1}%
	\addtocounter{footnote}{-1}%
	\endgroup
}

\begin{document}
\title{AutoLoc: Weakly-supervised Temporal Action Localization in Untrimmed Videos} 

\titlerunning{AutoLoc: Weakly-supervised Temporal Action Localization}
%
\author{Zheng Shou\inst{1} \and
Hang Gao\inst{1} \and
Lei Zhang\inst{2} \and
Kazuyuki Miyazawa\inst{3} \and
Shih-Fu Chang\inst{1}}
%
\authorrunning{Z. Shou, H. Gao, L. Zhang, K. Miyazawa, S.-F. Chang}
%

\institute{Columbia University, New York, NY, USA \and Microsoft Research, Redmond, Washington, USA \and Mitsubishi Electric, Japan}
\maketitle              
\begin{abstract}
	Temporal Action Localization (TAL) in untrimmed video is important for many applications. But it is very expensive to annotate the segment-level ground truth (action class and temporal boundary). This raises the interest of addressing TAL with weak supervision, namely only video-level annotations are available during training). However, the state-of-the-art weakly-supervised TAL methods only focus on generating good Class Activation Sequence (CAS) over time but conduct simple thresholding on CAS to localize actions. In this paper, we first develop a novel weakly-supervised TAL framework called AutoLoc to directly predict the temporal boundary of each action instance. We propose a novel Outer-Inner-Contrastive (OIC) loss to automatically discover the needed segment-level supervision for training such a boundary predictor. Our method achieves dramatically improved performance: under the IoU threshold 0.5, our method improves mAP on THUMOS'14 from 13.7\% to 21.2\% and mAP on ActivityNet from 7.4\% to 27.3\%. It is also very encouraging to see that our weakly-supervised method achieves comparable results with some fully-supervised methods.

	\keywords{Temporal Action Localization; Weak Supervision; Outer-Inner-Contrastive; Class Activation Sequence}
\end{abstract}

\blfootnote{Source code and models can be found at \color{blue}{https://github.com/zhengshou/AutoLoc}}

\section{Introduction}

Impressive improvement has been made in the past two years to address \textbf{Temporal Action Localization (TAL)} in untrimmed videos
\cite{fast_temporal_activity_cvpr16,victor_eccv16,Richard_2016_CVPR,stanford_cvpr16,yuan_cvpr16,scnn_shou_wang_chang_cvpr16,cdc_zheng_cvpr17,lin2017single,heilbron2017scc,zhao2017temporal,iccv17_tap,xu2017r,dave2017predictive,dai2017temporal,bmvc17_tad,buch2017sst,sstad_buch_bmvc17,yuan2017temporal,sigurdsson2017asynchronous}.
These methods were proposed for the \textbf{fully-supervised} setting: the model training requires the full annotation of the ground truth temporal boundary (start time and end time) for each action instance.
However, untrimmed videos are usually very long with substantial background content in time.
Therefore, manually annotating temporal boundaries for a new large-scale dataset is very expensive and time-consuming \cite{zhao2017slac}, and thus might prohibit applying the fully-supervised methods to the new domains that lack enough training data with full annotations.

\begin{figure}[t]
	\centering
	\includegraphics[width=\textwidth]{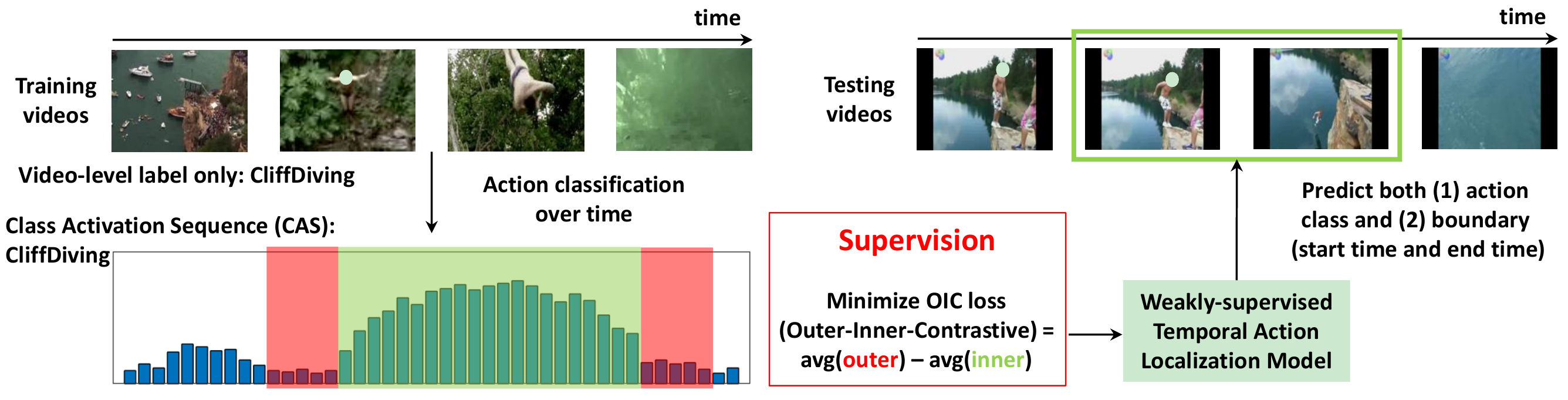}
	\caption{
		We study the weakly-supervised temporal action localization problem: during training we only have videos with the video-level labels, but during testing we need to predict both (1) the action class and (2) the temporal boundary of each action instance.
		In order to obtain the segment-level supervision for training the action localization model to predict the boundary directly, we design a novel \textbf{Outer-Inner-Contrastive (OIC)} loss based on the action Class Activation Sequence.
		We denote the predicted action segment boundary as the \textit{inner boundary}.
		The \textit{outer boundary} is obtained by extending the inner boundary to include its surrounding area.
		A desirable boundary prediction should have high activations in the inner \textcolor{green}{\textbf{\textbf{green}}} area but low activations in the outer \textcolor{red}{\textbf{\textbf{red}}} area.
		Consequently, the OIC loss can be used to approximately determine the needed segment-level supervision for training the localization model}
	\label{fig:intro}
\end{figure}

This motivates us to develop TAL methods that require significantly fewer ground truth annotations for training.
As illustrated in Fig.~\ref{fig:intro}, in this paper we focus on the following scenario: during training, we only have the video-level labels, which are much easier to collect , compared to the boundary annotations; during testing, we still aim to predict both (1) the action class and (2) the temporal boundary (i.e. start time and end time) of each action instance.
We refer this scenario as the \textbf{weakly-supervised} setting that this paper works on.

Recently, a few methods have been proposed to tackle TAL in such a weakly-supervised setting.
UntrimmedNet \cite{wang2017untrimmednets} and Hide-and-Seek \cite{singh2017hide}
achieve the state-of-the-art performances and carry out the localization in a similar manner.
Given a training video, several segments are randomly sampled and are fed into a network together to yield a video-level class prediction.
During testing, the trained network is slided over time to produce the classification score sequence of being each action over time.
The score sequence is similar to the Class Activation Map in \cite{zhou2016learning} but just has one dimension, and thus we refer it as \textbf{Class Activation Sequence (CAS)}.
Finally a simple thresholding method is applied on the CAS to localize each action instance in terms of the start time and the end time.

However, performing localization via thresholding in general may not be robust to noises in CAS: sometimes there are a few dips of low activations within an interval of high activations, using a large threshold might over-segment one action instance into several segments; but using a small threshold might include too many irrelevant backgrounds preceding and succeeding the action instance.
One possible solution is improving the quality of CAS.
Alternatively, instead of thresholding, many fully-supervised TAL methods detect action instances at the segment-level directly \cite{scnn_shou_wang_chang_cvpr16,sstad_buch_bmvc17}.
Some works further employ boundary regression models to learn to predict more accurate boundaries \cite{lin2017single,iccv17_tap,xu2017r,bmvc17_tad}. 
Thus, we design a framework called \textbf{AutoLoc} which can conduct direct boundary prediction via predicting the center location and the duration of each action instance.

But how to train the boundary prediction model without ground truth boundary annotations still remains unsolved.
To address this challenge, we propose a novel \textbf{Outer-Inner-Contrastive (OIC)} loss to provide the needed segment-level supervision for training the boundary prediction model.
Given the CAS of being the ground truth action, we denote the \textit{inner boundary} as the boundary of a predicted action instance and we inflate the inner boundary slightly to obtain the \textit{outer boundary}.
As illustrated in Fig.~\ref{fig:intro}, we propose an OIC loss as the average activation in the outer \textcolor{red}{\textbf{red}} area minus the average activation in the inner \textcolor{green}{\textbf{green}} area.
By minimizing the OIC loss to find the area of high inner activations but low outer activations, we can make desirable localization of the salient interval on CAS, which is likely to be well-aligned with the ground truth segment.
Equipped with the OIC loss, AutoLoc can automatically discover the segment-level supervision from the video-level annotations for training the boundary prediction model.
In Sec.~\ref{exp}, we will experimentally compare with the state-of-the-art methods and also study several variants of our model.

In summary, we make three novel contributions in this paper:

(1) To the best of our knowledge, AutoLoc is the first weakly-supervised TAL framework that can directly predict the temporal boundary of each action instance with only the video-level annotations available during training, specifically addressing the localization task at the segment level.

(2) To enable the training of such a parametric boundary prediction model, we design a novel OIC loss to automatically discover the segment-level supervision  and we prove that the OIC loss is differentiable to the underlying boundary prediction model.

(3) We demonstrate the effectiveness of AutoLoc on two standard benchmarks.
AutoLoc significantly outperforms the state-of-the-art weakly-supervised TAL methods and even achieves results comparable to some fully-supervised methods that use the boundary annotations during training.
When the overlap IoU threshold is set to 0.5 during evaluation, our method improves mAP on THUMOS'14 from 13.7\% to 21.2\% (54.7\% relative gain) and improves mAP on ActivityNet from 7.4\% to 27.3\% (268.9\% relative gain).

\section{Related Works}

\subsubsection{Video Action Analysis}
Detailed reviews can be found in recent surveys \cite{survey1,survey2,survey3,survey4,survey5,survey6}.
Researchers have developed quite a few deep networks for video action analysis such as 3D ConvNets \cite{3dcnn,ji,tran2017convnet}, LSTM \cite{lrcn2014}, two-stream network \cite{Simonyan14b}, I3D \cite{Kinetics}, etc.
For example, Wang et al. proposed Temporal Segment Network \cite{TSN}, which employed the two-stream network to model the long-range temporal structure in video and served as an effective backbone network in various video analysis tasks such as recognition \cite{TSN}, localization \cite{zhao2017temporal}, weakly-supervised learning \cite{wang2017untrimmednets}.

\subsubsection{Temporal Action Localization with Weak Supervision and Full Supervision}
Several large-scale video datasets have been created for TAL such as Charades \cite{Charades1,Charades2}, ActivityNet \cite{caba2015activitynet}, THUMOS \cite{THUMOS14,THUMOS15}.
In order to obtain the ground truth temporal boundaries to provide full supervision for training the fully-supervised TAL models, substantial efforts are required for annotating each of such large-scale datasets.
Therefore, it is useful and important to develop TAL models that can be trained with weak supervision only.

Video-level annotation is one kind of weak supervision that can be more easily collected and thus is quite interesting to the community.
Sun et al. \cite{sssn_mm15} was the first to consider TAL with only the video-level annotations available during training and the authors discovered the additional supervision from web images.
Recently, 
Singh et al. designed Hide-and-Seek \cite{singh2017hide} to address the challenge that weakly-supervised detection methods usually focus on the most discriminative parts while neglect other relevant parts of the target instance. 
Wang et al. \cite{wang2017untrimmednets} proposed a framework called UntrimmedNet consisting of a classification module to perform action classification and a selection module to detect important temporal segments. 
These recent methods are effectively learning an action classification model during training in order to generate reasonably good Class Activation Sequence (CAS) over time. But in order to detect temporal boundaries, a simple thresholding is applied on the CAS during testing.
Therefore, although these methods can excel at the video-level action recognition, the performance of temporal localization still has large room for improvement.

However, the fully-supervised TAL methods (boundary annotations available during training) have gone beyond the simple thresholding method.
First, some researchers performed localization at segment-level: they first generated the candidate segments via sliding window or proposal methods, and then they classified each segment into certain actions \cite{scnn_shou_wang_chang_cvpr16,iccv17_tap,xu2017r,bmvc17_tad,buch2017sst}.
Motivated by the success of single-shot object detection method \cite{liu2016ssd,redmon2016yolo9000,redmon2016you}, Lin et al. \cite{lin2017single} removed the proposal stage and directly conducted TAL in a single-shot fashion to simultaneously predict temporal boundary and action class.
Second, direct boundary prediction via anchor generation and boundary regression has been adapted from object detection \cite{liu2016ssd,redmon2016yolo9000,redmon2016you,fasterrcnn,fastrcnn} to fully-supervised TAL recently and proven to be quite effective in detecting more accurate boundaries \cite{lin2017single,zhao2017temporal,iccv17_tap,xu2017r,bmvc17_tad}.
This motivates us to generalize segment-level localization and direct boundary prediction to weakly-supervised TAL: we develop AutoLoc to first generate anchor segments and then regress their boundaries to obtain the predicted segments;
in order to train the boundary regressors, we propose the OIC loss to provide the segment-level supervision.

\subsubsection{Weakly-supervised Deep Learning Methods}
Other types of weak supervision for action detection have also been explored in the past.
For instance, Huang et al. \cite{huang2016connectionist} and Richard et al. \cite{richard2017weakly} both utilized the order of actions as the supervision used during training.
Mettes et al. \cite{mettes2016spot} worked on the spatio-temporal action detection using only the point-level supervision for training.

Weakly-supervised deep learning methods have been also widely studied in other vision tasks such as object detection \cite{zhou2016learning,zhu2017soft,shi2017weakly,kim2017two,jie2017deep,durand2017wildcat,peng2017multiple,zhang2016top,sun2016pronet,bilen2016weakly,kantorov2016contextlocnet,gudi2017object}, semantic segmentation \cite{khoreva2017simple,hong2017weakly,papandreou2015weakly,bearman2016s}, video captioning \cite{shen2017weakly}, visual relation detection \cite{zhang2017ppr}, etc.
As a counterpart of the weakly-supervised video TAL, the weakly-supervised image object detection has been significantly improved via combining Multiple Instance Learning (MIL) \cite{dietterich1997solving} and deep networks \cite{shi2017weakly,jie2017deep,peng2017multiple,bilen2016weakly,kantorov2016contextlocnet}: built upon Fast-RCNN \cite{fastrcnn}, these methods first generated candidate proposals beforehand; then they employed deep networks to classify each proposal and the scores from all proposals were fused together to obtain one label prediction for the whole image to be compared with the image-level label.
One of such MIL-based deep networks is ContextLocNet \cite{kantorov2016contextlocnet}, which further inflated the prediction box to obtain its outer box to take into account the contextual information. 
Our work bypasses the costly proposal generation and predicts the boundaries from raw input videos in a single-shot fashion.
Although we focus on video TAL in this paper, it would be also interesting to adapt our method for image object detection in the future.



\section{Outer-Inner-Contrastive Loss} \label{OIC}

In this Section, we formulate how to compute the proposed OIC loss during the network forward pass of AutoLoc and prove that the OIC loss is differentiable to the underlying boundary prediction model during the backward pass.
The whole pipeline and details of AutoLoc will be presented in Sec.~\ref{network}.

\subsection{Forward} \label{OIC:forward}

\begin{figure}[t]
	\centering
	\includegraphics[width=\textwidth]{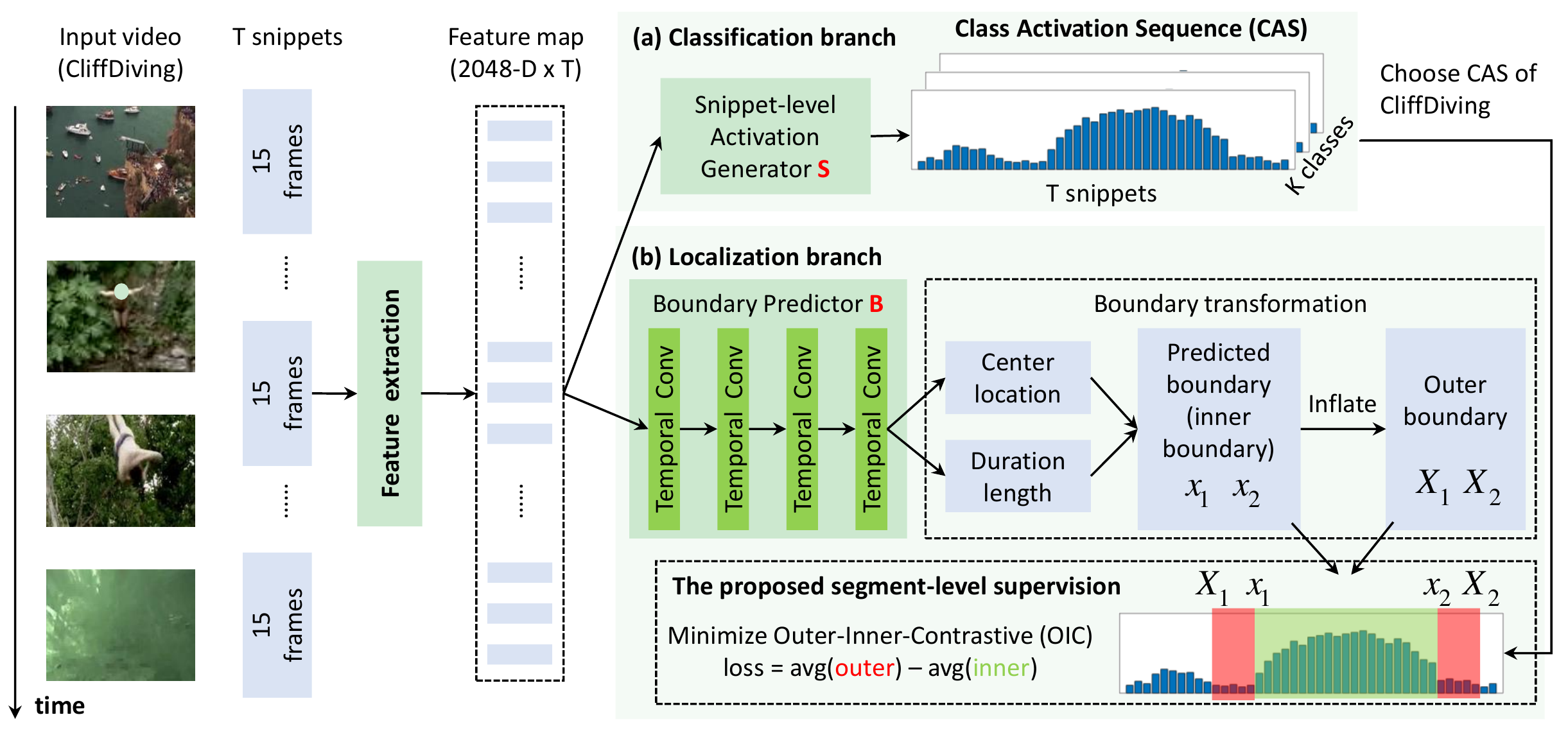}
	\caption{The network architecture of AutoLoc. Given an input video during training, the video is chunked into $T$ 15-frames-long snippets without overlap. We extract a feature vector of $D$ dimension for each snippet. On  top of the features, AutoLoc slides two separate branches over time:
		one classification branch for predicting the action scores of each snippet to obtain the \textbf{Class Activation Sequence (CAS)};
		one localization branch for directly predicting the true action boundary which is denoted as the inner boundary and is inflated to obtain the outer boundary.
		Based on the CAS of the ground truth video-level action, we can compute the Outer-Inner-Contrastive loss (\textit{the average activation in the outer \textcolor{red}{red} area minus the average activation in the inner \textcolor{green}{green} area}) to provide the needed segment-level supervision for training the boundary predictor}
	\label{fig:framework}
\end{figure}

As illustrated by the bottom-right part in Fig.~\ref{fig:framework}, for each predicted segment $\phi$, we can compute its OIC loss.
Each predicted segment $\phi$ consists of the action/inner boundary $\left[ {{x_1},{x_2}} \right]$, the inflated outer boundary $\left[ {{X_1},{X_2}} \right]$, and the action class $k$.
These boundaries are at the snippet-level granularity (for example, boundary $x=1$ corresponds to the location of the $1$-st snippet).
In order to fetch the corresponding snippet-level activation on the CAS, we round each boundary of continuous value to its nearest integer (i.e. the location of the nearest snippet).
We denote the class activation at the $x$-th snippet on the CAS of action $k$ as ${f_k}\left( x \right)$.
The OIC loss of the prediction $\phi$ is defined as the average activation ${A_o}\left( \phi  \right)$ in the outer area minus the average activation ${A_i}\left( \phi  \right)$ in the inner area:
\begin{equation}
{{\cal L}_{\rm{OIC}}}\left( \phi  \right) = {A_o}\left( \phi  \right) - {A_i}\left( \phi  \right) = \underbrace {\frac{{\int\limits_{{X_1}}^{{X_2}} {{f_k}\left( u \right)du}  - \int\limits_{{x_1}}^{{x_2}} {{f_k}\left( u \right)du} }}{{\left( {{X_2} - {X_1} + 1} \right) - \left( {{x_2} - {x_1} + 1} \right)}}}_{{A_o}\left( \phi  \right)} - \underbrace {\frac{{\int\limits_{{x_1}}^{{x_2}} {{f_k}\left( u \right)du} }}{{\left( {{x_2} - {x_1} + 1} \right)}}}_{{A_i}\left( \phi  \right)}.
\end{equation}
During training, we set $k$ to the ground truth action and we minimize ${{\cal L}_{\rm{OIC}}}\left( \phi  \right)$ to encourage high activations inside and penalize high activations outside.

\subsection{Backward} \label{OIC:backward}
We prove that the OIC loss is differentiable to the inner and outer boundaries.
Therefore, the supervision discovered by the OIC loss can be back-propagated to the underlying boundary prediction model.
Detailed derivation can be found in the supplementary material.
The gradients corresponding to the predicted segment $\phi$ w.r.t its inner boundary $\left[ {{x_1},{x_2}} \right]$ are as follows:
\begin{align}
\frac{{\partial {{\cal L}_{\rm{OIC}}}\left( \phi  \right)}}{{\partial {x_1}}} = \underbrace {\frac{{{f_k}\left( {{x_1}} \right) - {A_o}\left( \phi  \right)}}{{\left( {{X_2} - {X_1} + 1} \right) - \left( {{x_2} - {x_1} + 1} \right)}}}_{\frac{{\partial {A_o}\left( \phi  \right)}}{{\partial {x_1}}}} - \underbrace {\frac{{{A_i}\left( \phi  \right) - {f_k}\left( {{x_1}} \right)}}{{\left( {{x_2} - {x_1} + 1} \right)}}}_{\frac{{\partial {A_i}\left( \phi  \right)}}{{\partial {x_1}}}} \label{eq:x1};\\
\frac{{\partial {{\cal L}_{\rm{OIC}}}\left( \phi  \right)}}{{\partial {x_2}}} = \underbrace {\frac{{{A_o}\left( \phi  \right) - {f_k}\left( {{x_2}} \right)}}{{\left( {{X_2} - {X_1} + 1} \right) - \left( {{x_2} - {x_1} + 1} \right)}}}_{\frac{{\partial {A_o}\left( \phi  \right)}}{{\partial {x_2}}}} - \underbrace {\frac{{{f_k}\left( {{x_2}} \right) - {A_i}\left( \phi  \right)}}{{\left( {{x_2} - {x_1} + 1} \right)}}}_{\frac{{\partial {A_i}\left( \phi  \right)}}{{\partial {x_2}}}}
.
\end{align}
The gradients corresponding to the predicted segment $\phi$ w.r.t its outer boundary $\left[ {{X_1},{X_2}} \right]$ are as follows:
\begin{align}
\label{eq:X1}
\frac{{\partial {{\cal L}_{\rm{OIC}}}\left( \phi  \right)}}{{\partial {X_1}}} = \frac{{\partial {A_o}\left( \phi  \right)}}{{\partial {X_1}}} = \frac{{{A_o}\left( \phi  \right) - {f_k}\left( {{X_1}} \right)}}{{\left( {{X_2} - {X_1} + 1} \right) - \left( {{x_2} - {x_1} + 1} \right)}};\\
\frac{{\partial {{\cal L}_{\rm{OIC}}}\left( \phi  \right)}}{{\partial {X_2}}} = \frac{{\partial {A_o}\left( \phi  \right)}}{{\partial {X_2}}} = \frac{{{f_k}\left( {{X_2}} \right) - {A_o}\left( \phi  \right)}}{{\left( {{X_2} - {X_1} + 1} \right) - \left( {{x_2} - {x_1} + 1} \right)}}
.
\end{align}

Note that these gradients indeed have the physical meanings about how to adjust the boundaries.
For example, in Equation~\ref{eq:x1}, ${\frac{{\partial {A_i}\left( \phi  \right)}}{{\partial {x_1}}}}$ represents how much the average inner activation ${{A_i}\left( \phi  \right)}$ is higher than the activation ${{f_k}\left( {{x_1}} \right)}$ at the inner left boundary $x_1$.
If the average inner activation is much higher than the activation at the inner left boundary $x_1$, $x_1$ is likely to belong to the background and thus we would like to move $x_1$ in the positive (right) direction.
Similarly, ${\frac{{\partial {A_o}\left( \phi  \right)}}{{\partial {x_1}}}}$ represents how much the activation at the inner left boundary $x_1$ is higher than the average outer activation.
$\frac{{\partial {{\cal L}_{\rm{OIC}}}\left( \phi  \right)}}{{\partial {x_1}}}$ is the adversarial outcome of ${\frac{{\partial {A_o}\left( \phi  \right)}}{{\partial {x_1}}}}$ and ${\frac{{\partial {A_i}\left( \phi  \right)}}{{\partial {x_1}}}}$.
Consequently, $\frac{{\partial {{\cal L}_{\rm{OIC}}}\left( \phi  \right)}}{{\partial {x_1}}}$ indicates how the model wants to adjust the inner left boundary $x_1$ eventually:
if $\frac{{\partial {{\cal L}_{\rm{OIC}}}\left( \phi  \right)}}{{\partial {x_1}}} < 0$, $x_1$ moves in the positive (right) direction; if $\frac{{\partial {{\cal L}_{\rm{OIC}}}\left( \phi  \right)}}{{\partial {x_1}}} > 0$, $x_1$ moves in the negative (left) direction.

\section{AutoLoc} \label{network}

In this Section, we walk through the pipeline of AutoLoc as illustrated in Fig.~\ref{fig:framework}. 
The training and testing pipelines are very similar in AutoLoc.
So we only explicitly distinguish the training and testing pipelines when any difference appears.

\subsection{Input Data Preparation and Feature Extraction}

Each input data sample fed into AutoLoc is one single untrimmed video.
Following UntrimmedNet \cite{wang2017untrimmednets}, for each input video, we first divide it into 15-frames-long snippets without overlap and extract feature for each snippet individually.

In particular, Temporal Segment Network (TSN) \cite{TSN} is a state-of-the-art two-stream network for video analysis.
UntrimmedNet \cite{wang2017untrimmednets} has been proven to be effective in training TSN classifier with only the video-level labels.
Therefore, we first train an UntrimmedNet network (the soft version) in advance and then use the trained network as our backbone for feature extraction.

This backbone network consists of one spatial stream accepting RGB input and one temporal stream accepting Optical Flow input.
For each stream, we employ the Inception network architecture with Batch Normalization \cite{ioffe2015batch} and extract the 1024-dimensional feature at the $\tt global\_pool$ layer.
Finally, for each snippet, we concatenate the extracted spatial feature and temporal feature into one feature vector of 2048 dimensions.
For each input video of $T$ snippets in total, we obtain a feature map of shape 2048 (channels) by $T$ (snippets).

\subsection{Classification Branch}

The goal of the classification branch is to obtain the Class Activation Sequence (CAS).
We build our Activation Generator \textbf{S} based on UntrimmedNet.
On top of the $\tt global\_pool$ layer, UntrimmedNet attaches one Fully Connected (FC) layer of $K$ nodes to classify each snippet into $K$ action categories and also attaches another Fully Connected (FC) layer of just 1 node to predict the attention score (importance) for each snippet.
The corresponding scores from the spatial stream and the temporal stream are averaged to obtain the final score.
For each video, we use these two FC layers in the UntrimmedNet that are trained beforehand to respectively extract a classification score sequence of shape $K$ (actions) by $T$ (snippets) and an attention score sequence of $T$ dimensions.
For each snippet, we set its classification scores of all classes to 0 when its attention score is lower than the threshold (7 is chosen via grid search on the THUMOS'14 training set and also works well on ActivityNet);
then we regard such a gated classification score as the activation, which ranges within [0, 1].
Finally, for each video, we obtain its CAS of shape $K$ (actions) by $T$ (snippets).

\subsection{Localization Branch} \label{localization}

\subsubsection{Overview.}
The goal of the localization branch is to learn a parametric model for predicting the segment boundary directly.
Recent fully-supervised TAL methods \cite{lin2017single,zhao2017temporal,iccv17_tap,xu2017r,bmvc17_tad} have shown the effectiveness of regressing anchors for direct boundary prediction:
the anchor is a hypothesis of the possible segment;
the predicted boundary is obtained by respectively regressing (1) the center location and (2) the temporal length of the anchor segment;
multi-anchor mechanism is used to cover the possible segments of different temporal scales.
Therefore, we design a localization network \textbf{B} to look at each temporal position on the feature map and output the needed two boundary regression values for each anchor.
Then we regress the anchors using these regression values to obtain the predicted action boundaries (inner boundaries) and inflate the inner boundaries to obtain the outer boundaries.
Finally, based on the CAS, we introduce an OIC layer equipped with the OIC loss to generate the final segment predictions.

\subsubsection{Network Architecture of the Localization Network \textbf{B}}
Given an input video,  its feature map of shape 2048 channels by $T$ snippets
is fed into \textbf{B}.
\textbf{B} first stacks 3 same temporal convolutional layers, which slide convolutional filters over time.
Each temporal convolutional layer has 128 filters, which all have kernel size 3 in time with stride 1 and padding 1.
Each temporal convolutional layer is followed by one Batch Normalization layer and one ReLU layer.

Finally, \textbf{B} adds one more temporal convolutional layer $\tt pred$ to output the boundary regression values.
Filters in $\tt pred$ have kernel size 3 in time with stride 1 and padding 1.
Similar to YOLO \cite{redmon2016yolo9000,redmon2016you}, the boundary predicted by \textbf{B} is designed to be class-agnostic.
This allows us to learn a generic boundary predictor, which may be used for generating action proposals for unseen actions in the future.
Consequently, the total number of filters in $\tt pred$ is $2M$, where $M$ is the number of anchor scales.
For each anchor, \textbf{B} predicts two boundary regression values: (1) $t_x$ indicating how to shift the center location of the anchor and (2) $t_w$ indicating how to scale the length of the anchor.

\subsubsection{Details of the Boundary Transformation}

\begin{figure}[t]
	\centering
	\includegraphics[width=\textwidth]{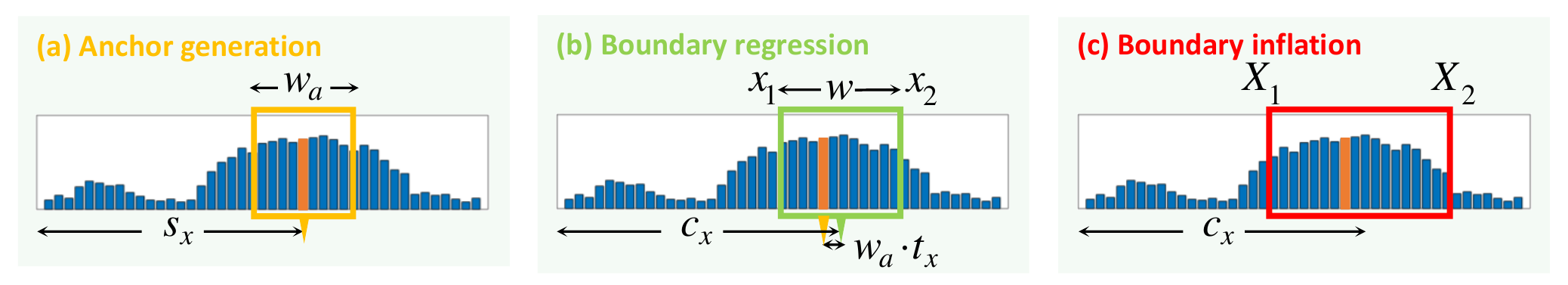}
	\caption{Illustration of the boundary prediction procedure which consists of three steps sequentially: (1) \textbf{anchor generation} to obtain the boundary hypothesis; (2) \textbf{boundary regression} to obtain the predicted boundary of the action segment (denoted as the inner boundary); (3) \textbf{boundary inflation} to obtain the outer boundary. The score sequence is CAS and the orange score bar indicates the temporal position that the boundary predictor currently looks at}
	\label{fig:boundary}
\end{figure}

Since each temporal position on the feature map and each temporal position on the CAS both correspond to the same location of an input snippet, we make boundary predictions at the snippet-level granularity.
We outline the boundary prediction procedure in Fig.~\ref{fig:boundary}.

{\it Anchor generation.}
At the temporal position $s_x$ on the feature map, we generate a hypothesized segment (anchor) of length $w_a$. In practice, we use multi-scale anchors.
We determine their scales according to the typical time duration range of segments in each specific dataset.

{\it Boundary regression.}
As aforementioned, for each anchor at the temporal position $s_x$, \textbf{B} predicts two boundary regression values $t_x$ and $t_w$.
We can obtain the predicted segment via regressing the center location 
${c_x} = {s_x} + {w_a} \cdot {t_x}$
and the temporal length
$w = {w_a} \cdot \exp \left( {{t_w}} \right)$.
We denote the boundary of this predicted segment as the inner boundary, which can be computed by ${x_1} = {c_x} - w/2$ and ${x_2} = {c_x} + w/2$.
Furthermore, we clip the predicted boundary $x_1$ and $x_2$ to fit into the range of the whole video.
More details about clipping can be found in the supplementary material.

{\it Boundary inflation.}
A ground truth segment usually exhibits relatively higher activations on CAS within the inner area $\left[ {{x_1},{x_2}} \right]$ compared to the contextual area preceding $x_1$ and succeeding $x_2$.
Therefore, we inflate the inner boundary by a ratio $\alpha$ to obtain the corresponding outer boundary 
${X_1} = {x_1} - w \cdot \alpha $
and 
${X_2} = {x_2} + w \cdot \alpha $.
As discussed in the supplementary material, setting $\alpha $ to 0.25 is a good choice.

\subsubsection{The OIC layer for Obtaining the Final Predictions}
\label{postprocessing}

Finally, we introduce an OIC layer which uses the OIC loss to measure how likely each segment contains actions and then removes the segments that are not likely to contain actions.
During testing, this OIC layer outputs a set of predicted segments.
During training, this OIC layer further computes the total OIC loss and back-propagates the gradients to the underlying boundary prediction model.

Concretely, given an input video, the classification branch generates its CAS and the localization branch predicts the candidate class-agnostic segments.
Note that since all temporal convolutional layers in \textbf{B} slide over time with stride 1, the set of segments predicted at each temporal position on the feature map and the activations at each temporal position on the CAS are paired, corresponding to the same input snippet.
Thus at the temporal position of each snippet, \textbf{B} has predicted $M$ class-agnostic anchor segments.
Then for each action, we iteratively go through the following steps on the CAS to obtain the final class-specific segment predictions.
Note that during training we consider only the ground truth actions while during testing we consider all actions.
If a temporal position has the activation lower than 0.1 on the CAS, we discard all the predictions corresponding to this temporal position.
For each of the remaining positions, among its $M$ anchor segment predictions, we only keep the one with the lowest OIC loss which means selecting the anchor of the most likely scale.
Finally, for all the kept segment predictions, we remove the segment predictions with the OIC loss higher than -0.3.
We perform Non-Maximum Suppression (NMS) over all segment predictions with overlap IoU threshold 0.4.
All these thresholds are chosen by grid search on the THUMOS'14 training set and also work well on ActivityNet.
Alg. 1 in the supplementary material summarizes the above steps.


During training, the total loss is the summation of the OIC loss generated by each kept segment predictions.
We can compute the gradients triggered by each kept segment prediction according to Sec.~\ref{OIC:backward} and then accumulate them together to update the underlying boundary predictor \textbf{B}.
During testing, all the kept segment predictions are outputted as our final segment predictions.
Each segment prediction consists of (1) the predicted action class, (2) the confidence score which is set to 1 minus its OIC loss, and (3) the start time and the end time obtained by converting the inner boundary [$x_1$, $x_2$] from the snippet-level granularity (continuous value before rounding to its nearest integer) to time.

\section{Experiments} \label{exp}

In this section, we first introduce two standard benchmarks and the corresponding evaluation metrics.
Note that during training, we only use the video-level labels; during testing, we use the ground truth segments with boundary annotations for evaluating the performance of temporal action localization.
We compare our method with the state-of-the-art methods and then conduct some ablation studies to investigate different variants of our method.

\subsection{Datasets and Evaluation}

\subsubsection{THUMOS'14 \cite{THUMOS14}}
The temporal action localization task in THUMOS'14 contains 20 actions. Its validation set has 200 untrimmed videos. Each video contains at least one action. We use these 200 videos in the validation set for training. The trained model is tested on the test set which contains 213 videos.

\subsubsection{ActivityNet v1.2 \cite{caba2015activitynet}}
To facilitate comparisons, we follow Wang et al. \cite{wang2017untrimmednets} to use the ActivityNet release version 1.2 which covers 100 activity classes. The training set has 4,819 videos and the validation set has 2,383 videos.
We train on the training set and test on the validation set.

\subsubsection{Evaluation Metrics}
Given the testing videos, the system outputs a rank list of action segment predictions. Each prediction contains the action class, the starting time and the ending time, and the confidence score.
We follow the conventions \cite{THUMOS14,activitynet} to evaluate mean Average Precision (mAP).
Each prediction is regarded as correct only when (1) the predicted class is correct and (2) its temporal overlap IoU with the ground truth segment exceeds the evaluation threshold.
We do not allow duplicate detections for the same ground truth segment.

\subsection{Implementation Details}
We implement our AutoLoc using Caffe \cite{caffe}.
We use the stochastic gradient descent algorithm to train AutoLoc.
Through the experimental studies, we find that the training process can converge quickly on both THUMOS'14 and ActivityNet datasets after 1 training epoch.
Following Faster R-CNN \cite{fasterrcnn}, during each mini-batch, we process one whole untrimmed video. 
The learning rate is initially set to 0.001 and is reduced by one order of  magnitude for every 200 iterations on THUMOS'14 and for every 500 iterations on ActivityNet.
We set the weight decay to 0.0005.
We choose anchors of the snippet-level length 1, 2, 4, 8, 16, 32 for THUMOS'14 and 16, 32, 64, 128, 256, 512 for ActivityNet.
We use CUDA 8.0 and cuDNN v5. We use one single NVIDIA GeForce GTX TITAN X GPU. 


\begin{table}[t]
	\begin{center}
		\caption{Comparisons with the state-of-the-art methods in terms of temporal localization mAP (\%) under different IoU thresholds on THUMOS'14 test set. Weak supervision means training with the video-level labels only. Full supervision indicates that the segment-level boundary annotations are used during training}
		\label{table:res_th}
		\begin{tabular}{c|c|ccccc}
			Supervision & IoU threshold     & 0.3           & 0.4           & 0.5           & 0.6           & 0.7          \\ \hline
			Full        & Karaman et al. \cite{th3} &  0.5 &  0.3 &  0.2 &  0.2 &  0.1   \\
			Full        & Wang et al. \cite{th2} &  14.6 &  12.1 &  8.5 & 4.7 &  1.5   \\
			Full        & Heilbron et al. \cite{fast_temporal_activity_cvpr16} &  - &  - &  13.5 &  - &  -   \\
			Full        & Escorcia et al. \cite{victor_eccv16} &  - &  - &  13.9 &  - &  -  \\
			Full        & Oneata et al. \cite{th1}  &  28.8 &  21.8 &  15.0  &  8.5 &  3.2 \\
			Full        & Richard and Gall \cite{Richard_2016_CVPR} &  30.0 &  23.2 &  15.2  &  - &  -  \\
			Full        & Yeung et al. \cite{stanford_cvpr16} &  36.0 &  26.4  &  17.1 &  - &  -  \\
			Full        & Yuan et al. \cite{yuan_cvpr16} &  33.6 &  26.1 &  18.8  &  - & -   \\
			Full        & Yuan et al.  \cite{yuan2017temporal} &  36.5 &  27.8 & 17.8 &  - &  -  \\ 
			Full        & S-CNN  \cite{scnn_shou_wang_chang_cvpr16} &  36.3 &  28.7 & 19.0 &  10.3 &  5.3  \\ 
			Full        & SST \cite{buch2017sst}          & 37.8          & -          & 23.0          & -          & -          \\
			Full        & CDC \cite{cdc_zheng_cvpr17}          & 40.1          & 29.4          & 23.3          & 13.1          & 7.9          \\
			Full        & Dai et al. \cite{dai2017temporal}          & -          & 33.3          & 25.6          & 15.9       & 9.0 \\
			Full        & SSAD \cite{lin2017single}          & 43.0          & 35.0          & 24.6          & -       & - \\
			Full        & TURN TAP \cite{iccv17_tap}          & 44.1          & 34.9          & 25.6          & -       & - \\
			Full        & R-C3D \cite{xu2017r}          & 44.7          & 35.6          & 28.9          & -       & - \\
			Full        & SS-TAD \cite{sstad_buch_bmvc17}          & 45.7          & -          & 29.2          & -       & 9.6 \\
			Full        & Gao et al. \cite{bmvc17_tad}          & 50.1          & \textbf{41.3}          & \textbf{31.0}          & 19.1       & 9.9 \\
			Full        & SSN    \cite{zhao2017temporal}      & \textbf{51.9}          & 41.0          & 29.8          & \textbf{19.6}          & \textbf{10.7}         \\ \hline
			Weak        & Sun et al. \cite{sssn_mm15} & 8.5          & 5.2          & 4.4           & -             & -            \\
			Weak        & Hide-and-Seek \cite{singh2017hide} & 19.5          & 12.7          & 6.8           & -             & -            \\
			Weak        & Wang et al. \cite{wang2017untrimmednets} & 28.2          & 21.1          & 13.7          & -             & -            \\
			Weak        & Ours - AutoLoc       & \textbf{35.8} & \textbf{29.0} & \textbf{21.2} & \textbf{13.4} & \textbf{5.8}
		\end{tabular}
	\end{center}
\end{table}

\subsection{Comparisons with the State-of-the-art}

The results on THUMOS'14 are shown in Table~\ref{table:res_th}.
Our method significantly outperforms the state-of-the-art weakly-supervised TAL methods that are trained with the video-level labels only.
Regarding to the recent weakly-supervised TAL methods (i.e. Hide-and-Seek \cite{singh2017hide} and Wang et al. \cite{wang2017untrimmednets}),
although they can generate reasonably good CAS, TAL is done by applying simple thresholding on the CAS which might not robust be to noises in CAS.
Our method directly predicts the segment boundary with the contextual information taken into account.
Our method can even achieve better or comparable results to some fully-supervised methods (e.g. S-CNN \cite{scnn_shou_wang_chang_cvpr16}) that are trained with the segment-level boundary annotations.
The results of SSN \cite{zhao2017temporal} correspond to the model of the same backbone network architecture as ours.

The results on ActivityNet v1.2 are shown in Table~\ref{table:res_an} and our method can achieve substantial improvements again.
Wang et al. \cite{wang2017untrimmednets} did not report temporal localization results on ActivityNet in their paper. But their trained models and source codes have been released online publicly and thus we can evaluate their results on ActivityNet as well.

\begin{table}
	\begin{center}
		\caption{Comparisons with the state-of-the-art methods in terms of temporal localization mAP (\%) under different IoU thresholds on ActivityNet v1.2 validation set. Weak supervision means training with the video-level labels only. Full supervision indicates that the segment-level boundary annotations are used during training}
		\label{table:res_an}
		\begin{tabular}{c|c|cccccccccc|c}
			Supervision & IoU threshold    & 0.5           & 0.55          & 0.6           & 0.65          & 0.7           & 0.75          & 0.8           & 0.85         & 0.9          & 0.95         & Avg           \\ \hline
			Full        & SSN     \cite{zhao2017temporal}     & 41.3          & 38.8          & 35.9          & 32.9          & 30.4          & 27.0          & 22.2          & 18.2         & 13.2         & 6.1          & 26.6          \\ \hline
			Weak        & Wang et al. \cite{wang2017untrimmednets} & 7.4           & 6.1           & 5.2           & 4.5           & 3.9           & 3.2           & 2.5           & 1.8          & 1.2          & 0.7          & 3.6           \\
			Weak        & Ours - AutoLoc      & \textbf{27.3} & \textbf{24.9} & \textbf{22.5} & \textbf{19.9} & \textbf{17.5} & \textbf{15.1} & \textbf{13.0} & \textbf{10.0} & \textbf{6.8} & \textbf{3.3} & \textbf{16.0}
		\end{tabular}
	\end{center}
\end{table}

\subsection{Discussions}

In this Section, we address several questions quantitatively to analyze our model.

\subsubsection{Q1: How Effective is the Proposed OIC Loss?}

In order to evaluate the effectiveness of the proposed OIC loss,
we enumerate all candidate segments at the snippet-level granularity (for example, a segment starting at the location of the 2-nd snippet and ending at the location of the 6-th snippet).
We leverage the OIC loss to measure how likely each segment contains actions and then select the most likely ones.
Concretely, for each segment, we compute its OIC loss of being each action.
Then we follow Sec.~\ref{postprocessing} to remove segments with high OIC loss and remove duplicate predictions via NMS.
We denote this approach as \textbf{OIC Selection}.
As shown in Table~\ref{table:res_discussion},
although not as good as AutoLoc, \textbf{OIC Selection} still significantly improves the state-of-the-art results \cite{wang2017untrimmednets}.
Because the OIC loss explicitly favors the segment which has high activations inside and low activations outside, and also such a segment of low OIC loss is usually well aligned to the ground truth segment. 
This confirms the effectiveness of the proposed OIC loss.

\subsubsection{Q2: How Important is Looking into the Contrast Between the Inner Area and the Outer Area?}

The core idea of the OIC loss is encouraging high activations in the inner area while penalizing high activations in the outer area.
We consider another variant that can also discover the segment-level supervision but does not model the contrast between inner and outer.
Specifically, we change the OIC loss in AutoLoc to \textbf{Inner Only Loss}, which only encourages high activations inside the segment but does not look into the contextual area.
The detailed formulation of how to compute the loss and gradients in the \textbf{Inner Only Loss} can be found in the supplementary material.
As shown in Table~\ref{table:res_discussion}, the performances drop a lot.
Consequently, when designing the loss for training the boundary predictor, it is very important and effective to take into account the contrast between the inner area and the outer area.

\begin{figure}[t]
	\centering
	\includegraphics[width=.8\textwidth]{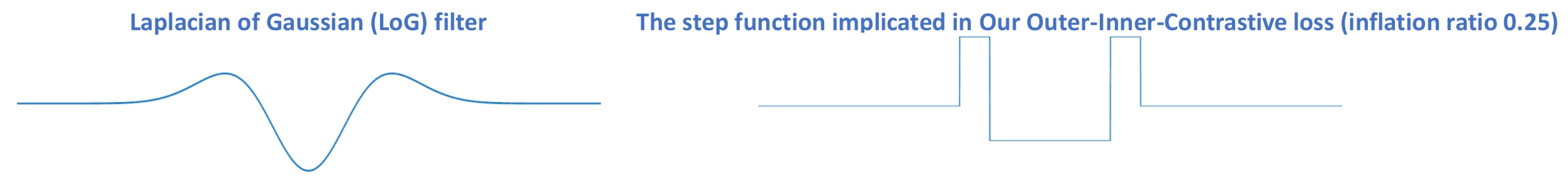}
	\caption{Illustration of the LoG filter and our OIC loss, which in effect is a step function}
	\label{fig:LoG}
\end{figure}

Notably, the idea of looking into the contrast between inner and outer is related to the usage of \textbf{Laplacian of Gaussian (LoG)} filter for blob detection \cite{lindeberg1998feature}.
The operation of computing the OIC loss is effectively convolving the CAS with a step function as shown in Fig.~\ref{fig:LoG}, which can be regarded as a variant of the LoG filter for the sake of easing the network training.
As proven in the supplementary material, the integral of the LoG filter and the integral of the step function are both zero on the range $(-\rm{Inf},+\rm{Inf})$.
Further, we approach the scale selection in blob detection by the multi-anchor mechanism and the boundary regression method.
Despite the simplicity of the OIC loss, it turns out to be quite effective in practice for localizing likely action segments.

\begin{table} [t]
	\begin{center}
		\caption{Temporal localization mAP (\%) under different IoU thresholds on ActivityNet v1.2 validation set. All approaches are trained with weak supervision}
		\label{table:res_discussion}
		\begin{tabular}{c|cccccccccc|c}
			IoU threshold    & 0.5           & 0.55          & 0.6           & 0.65          & 0.7           & 0.75          & 0.8           & 0.85         & 0.9          & 0.95         & Avg           \\ \hline
			Wang et al. \cite{wang2017untrimmednets} & 7.4           & 6.1           & 5.2           & 4.5           & 3.9           & 3.2           & 2.5           & 1.8          & 1.2          & 0.7          & 3.6           \\
			Ours - AutoLoc      & \textbf{27.3} & \textbf{24.9} & \textbf{22.5} & \textbf{19.9} & \textbf{17.5} & \textbf{15.1} & \textbf{13.0} & \textbf{10.0} & \textbf{6.8} & \textbf{3.3} & \textbf{16.0}\\ \hline
			Q1: OIC Selection & 15.8           & 13.7           & 11.9           & 10.3           & 8.8           & 7.5           & 6.4           & 5.1          & 3.6          & 2.2          & 8.5           \\ 
			Q2: Inner Only Loss & 4.6           & 3.7           & 2.7           & 1.9          & 1.3           & 0.9           & 0.5           & 0.2          & 0.1          & 0.0          & 1.6           \\ 
			Q3: Direct Optimization & 21.8           & 19.6           & 17.8           & 15.8          & 13.8           & 11.7           & 9.8           & 7.8          & 5.5          & 2.7          & 12.6                     
		\end{tabular}
	\end{center}
\end{table}

\subsubsection{Q3: What is the Advantage of Learning a Model on the Training Videos Compared to Directly Optimizing the Boundaries on the Testing Videos?}

AutoLoc trains a model on the training videos and then applies the trained model to perform inference on the testing videos.
Alternatively, without training the boundary predictor \textbf{B} on the training videos, we can directly train/optimize \textbf{B} from scratch on each testing video individually:
we follow the testing pipeline as described in Sec.~\ref{postprocessing} while we also conduct the back-propagation to update \textbf{B} to iteratively find likely segments on each testing video.
We refer this approach as \textbf{Direct Optimization}.
As shown in Table~\ref{table:res_discussion}, its performance is not bad, which confirms the effectiveness of the OIC loss again.
But it is still not as good as AutoLoc.
Because \textbf{Direct Optimization} optimizes the predicted boundaries according to the testing video's CAS, which may not be very accurate.
Eventually \textbf{Direct Optimization} overfits such an inaccurate CAS and thus results into imperfect boundary predictions.
In AutoLoc, \textbf{B} has been trained on multiple training videos and thus is robust to the noises in CAS.
Consequently, AutoLoc may still predicts good boundary even when the testing video's CAS is not perfect.
Furthermore, \textbf{Direct Optimization} requires optimizing the boundary predictions on the testing video until convergence and thus its testing speed is much slower than AutoLoc.
For example, on ActivityNet, \textbf{Direct Optimization} converges after 25 training iterations (25 forward passes and 25 backward passes).
However, AutoLoc directly applies the trained model to do inference on the testing video and thus requires only one forward pass during testing.

\section{Conclusion and Future Works}


In this paper, we have presented a novel weakly-supervised TAL framework to directly predict temporal boundary in a single-shot fashion and proposed a novel OIC loss to provide the needed segment-level supervision.
In the future, it would be interesting to extend AutoLoc for object detection in image.

\section{Acknowledgment}

We appreciate the support from Mitsubishi Electric for this project.

%
%
%

\clearpage

\section{Supplementary Materials}

\subsection{Visualization Examples} \label{sec:5}

In this section, we show two sets of experimental testing results on the THUMOS'14 test set.
We can observe that using a simple thresholding sometimes over-segments a whole action instance into two segments while using a simple thresholding sometimes mistakenly merges two consecutive action instances into one segment. But our AutoLoc method is robust to the noises in CAS and can localize the action instances correctly.

\begin{figure}
	\centering
	\includegraphics[width=.95\textwidth]{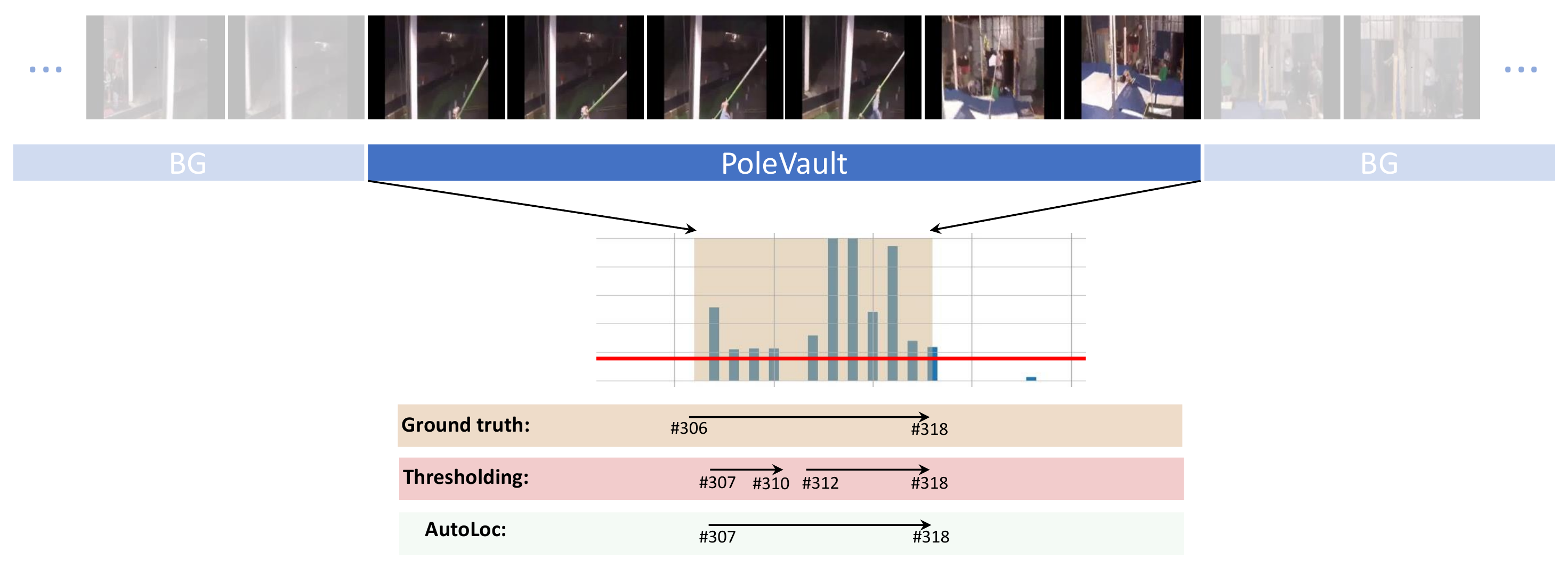}
	\caption{Visualization of the experimental testing results of the simple thresholding method and our AutoLoc method for a PoleVault action instance in the THUMOS'14 test set. The score sequence is the CAS of being PoleVault. Three bars in the bottom are respectively the ground truth segment, the segments detected by thresholding, and the segment detected by AutoLoc. The numbers shown with \# are snippet indexes}
	\label{fig:vis-1}
\end{figure}

As shown in Fig.~\ref{fig:vis-1}, the activations of being PoleVault in the first half part of the ground truth segment are low due to that the camera does not capture the full human body.
In this case, the threshold used in the simple thresholding method is relatively high so that it cuts the whole segment into two segments.
However, our AutoLoc model has been trained on multiple training videos to predict at the segment level.
Thus, our AutoLoc is robust to such noises in CAS and can predict the correct segment as a whole.

\begin{figure}
	\centering
	\includegraphics[width=\textwidth]{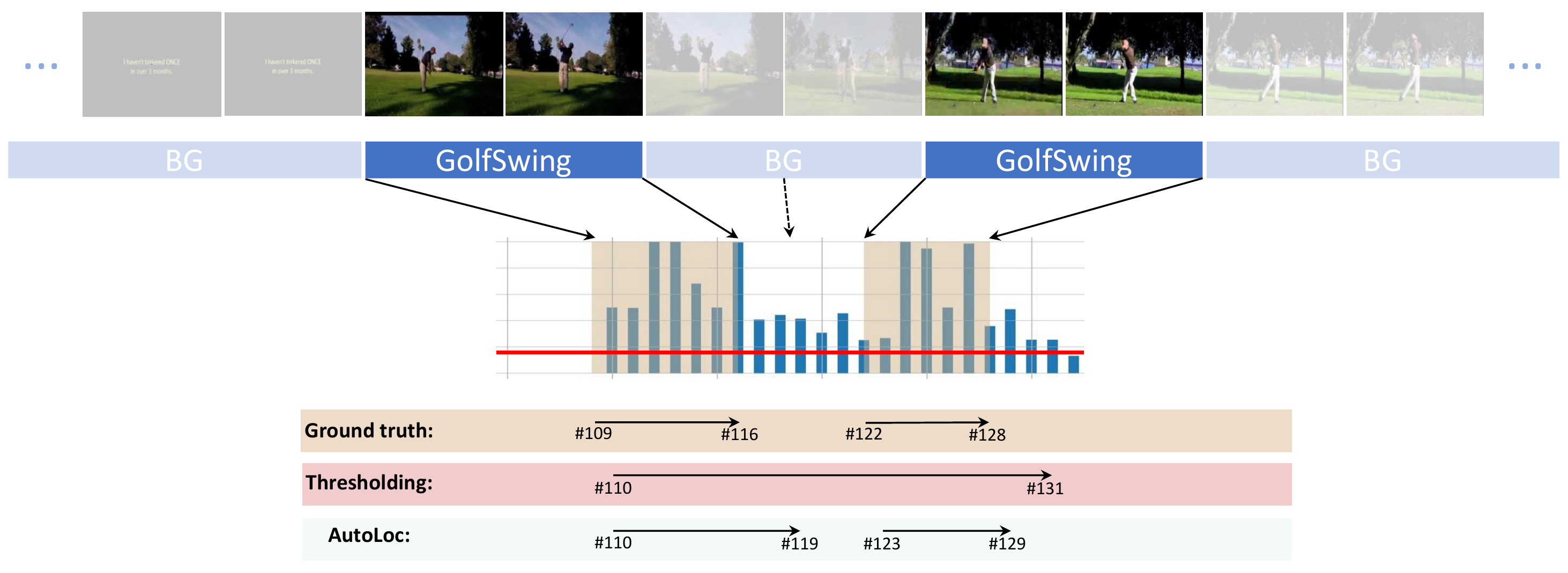}
	\caption{Visualization of the experimental testing results of the simple thresholding method and our AutoLoc method for two GolfSwing action instances in the THUMOS'14 test set.  The score sequence is the CAS of being GolfSwing. Three bars in the bottom are respectively the ground truth segments, and the segment detected by thresholding, the segments detected by AutoLoc. The numbers shown with \# are snippet indexes.}
	\label{fig:vis-2}
\end{figure}

Although the threshold used in the simple thresholding method is relatively high for the example shown in Fig.~\ref{fig:vis-1}, the same threshold is actually low in other cases such as the one shown in Fig.~\ref{fig:vis-2}.
In Fig.~\ref{fig:vis-2}, there are two GolfSwing instances happened consecutively. After the first instance ends and before the second instance starts, the scene of the first instance shades into the scene of the second instance, resulting into high
activations of being GolfSwing within this interval of the transition.
The thresholding method mistakenly detects these two instances as one whole  segment.
But our AutoLoc model can correctly detect the ending of the first instance and the beginning of the second instance and thus our AutoLoc can localize these two segments separately.
By the way, note that the ending boundary of the first instance predicted by AutoLoc is not very well-aligned with the annotated ground truth.
We notice that the corresponding video contents from the ground truth ending boundary to our detected ending boundary are that the person keeps still with his arms lifted.
It is actually kind of ambiguous to determine when the first GolfSwing instance ends precisely and thus the ending boundary predicted by our AutoLoc shall also be acceptable.

\subsection{Detailed Derivation of the OIC Back-propagation} \label{sec:1}

Here we present the detailed derivation about how to calculate the gradients of the OIC loss during back-propagation.

Each predicted segment $\phi$ consists of the action/inner boundary $\left[ {{x_1},{x_2}} \right]$, and the inflated outer boundary $\left[ {{X_1},{X_2}} \right]$.
These boundaries are at the snippet-level granularity (for example, boundary $x=1$ corresponds to the location of the $1$-st snippet).
We denote the class activation at the $x$-th snippet on the CAS of being the action $k$ as ${f_k}\left( x \right)$.
The OIC loss of the prediction $\phi$ (i.e. ${{\cal L}_{\rm{OIC}}}\left( \phi  \right)$) is defined as the average activation ${A_o}\left( \phi  \right)$ in the outer area minus the average activation ${A_i}\left( \phi  \right)$ in the inner area as follows:
\begin{align}
{A_o}\left( \phi  \right) & = \frac{{\int\limits_{{X_1}}^{{X_2}} {{f_k}\left( u \right)du}  - \int\limits_{{x_1}}^{{x_2}} {{f_k}\left( u \right)du} }}{{\left( {{X_2} - {X_1} + 1} \right) - \left( {{x_2} - {x_1} + 1} \right)}};\\
{A_i}\left( \phi  \right) & = \frac{{\int\limits_{{x_1}}^{{x_2}} {{f_k}\left( u \right)du} }}{{\left( {{x_2} - {x_1} + 1} \right)}};\\
{{\cal L}_{\rm{OIC}}}\left( \phi  \right) & = {A_o}\left( \phi  \right) - {A_i}\left( \phi  \right).
\end{align}

The gradient corresponding to the predicted segment $\phi$ w.r.t the left outer boundary $X_1$ of $\phi$ is as follows:
\begin{align}
\frac{{\partial {{\cal L}_{\rm{OIC}}}\left( \phi  \right)}}{{\partial {X_1}}} & = \frac{{\partial {A_o}\left( \phi  \right)}}{{\partial {X_1}}}\\
& = \frac{{\partial \frac{{\int\limits_{{X_1}}^{{X_2}} {{f_k}\left( u \right)du}  - \int\limits_{{x_1}}^{{x_2}} {{f_k}\left( u \right)du} }}{{\left( {{X_2} - {X_1} + 1} \right) - \left( {{x_2} - {x_1} + 1} \right)}}}}{{\partial {X_1}}}\\
& = \frac{{\frac{{\partial \int\limits_{{X_1}}^{{X_2}} {{f_k}\left( u \right)du}  - \int\limits_{{x_1}}^{{x_2}} {{f_k}\left( u \right)du} }}{{\partial {X_1}}} - {A_o}\left( \phi  \right)\frac{{\left( {{X_2} - {X_1} + 1} \right) - \left( {{x_2} - {x_1} + 1} \right)}}{{\partial {X_1}}}}}{{\left( {{X_2} - {X_1} + 1} \right) - \left( {{x_2} - {x_1} + 1} \right)}}\\
& = \frac{{\frac{{\partial \int\limits_{{X_1}}^{{X_2}} {{f_k}\left( u \right)du}  - \int\limits_{{x_1}}^{{x_2}} {{f_k}\left( u \right)du} }}{{\partial {X_1}}} + {A_o}\left( \phi  \right)}}{{\left( {{X_2} - {X_1} + 1} \right) - \left( {{x_2} - {x_1} + 1} \right)}}\\
& = \frac{{\frac{{\partial \int\limits_{{X_1}}^{{X_2}} {{f_k}\left( u \right)du} }}{{\partial {X_1}}} + {A_o}\left( \phi  \right)}}{{\left( {{X_2} - {X_1} + 1} \right) - \left( {{x_2} - {x_1} + 1} \right)}}\\
& = \frac{{{A_o}\left( \phi  \right) + {f_k}\left( {{X_2}} \right)\frac{{\partial {X_2}}}{{\partial {X_1}}} - {f_k}\left( {{X_1}} \right)\frac{{\partial {X_1}}}{{\partial {X_1}}}}}{{\left( {{X_2} - {X_1} + 1} \right) - \left( {{x_2} - {x_1} + 1} \right)}}\\
& = \frac{{{A_o}\left( \phi  \right) - {f_k}\left( {{X_1}} \right)}}{{\left( {{X_2} - {X_1} + 1} \right) - \left( {{x_2} - {x_1} + 1} \right)}}.
\end{align}

Likewise, the gradient corresponding to the predicted segment $\phi$ w.r.t the right outer boundary $X_2$ of $\phi$ can be computed as follows:
\begin{align}
\frac{{\partial {{\cal L}_{\rm{OIC}}}\left( \phi  \right)}}{{\partial {X_2}}} & = \frac{{\partial {A_o}\left( \phi  \right)}}{{\partial {X_2}}} \\
& = \frac{{{f_k}\left( {{X_2}} \right) - {A_o}\left( \phi  \right)}}{{\left( {{X_2} - {X_1} + 1} \right) - \left( {{x_2} - {x_1} + 1} \right)}}
.
\end{align}

The gradient corresponding to the predicted segment $\phi$ w.r.t the left inner boundary $x_1$ of $\phi$ is as follows:
\begin{align}
\frac{{\partial {{\cal L}_{\rm{OIC}}}\left( \phi  \right)}}{{\partial {x_1}}} & = \frac{{\partial {A_o}\left( \phi  \right)}}{{\partial {x_1}}} - \frac{{\partial {A_i}\left( \phi  \right)}}{{\partial {x_1}}};
\end{align}
\begin{align}
\frac{{\partial {A_o}\left( \phi  \right)}}{{\partial {x_1}}} &  = \frac{{\partial \frac{{\int\limits_{{X_1}}^{{X_2}} {{f_k}\left( u \right)du}  - \int\limits_{{x_1}}^{{x_2}} {{f_k}\left( u \right)du} }}{{\left( {{X_2} - {X_1} + 1} \right) - \left( {{x_2} - {x_1} + 1} \right)}}}}{{\partial {x_1}}}\\
& = \frac{{\frac{{\partial \int\limits_{{X_1}}^{{X_2}} {{f_k}\left( u \right)du}  - \int\limits_{{x_1}}^{{x_2}} {{f_k}\left( u \right)du} }}{{\partial {x_1}}} - {A_o}\left( \phi  \right)\frac{{\left( {{X_2} - {X_1} + 1} \right) - \left( {{x_2} - {x_1} + 1} \right)}}{{\partial {x_1}}}}}{{\left( {{X_2} - {X_1} + 1} \right) - \left( {{x_2} - {x_1} + 1} \right)}}\\
& = \frac{{\frac{{\partial \int\limits_{{X_1}}^{{X_2}} {{f_k}\left( u \right)du}  - \int\limits_{{x_1}}^{{x_2}} {{f_k}\left( u \right)du} }}{{\partial {x_1}}} - {A_o}\left( \phi  \right)}}{{\left( {{X_2} - {X_1} + 1} \right) - \left( {{x_2} - {x_1} + 1} \right)}}\\
& = \frac{{ - \frac{{\partial \int\limits_{{x_1}}^{{x_2}} {{f_k}\left( u \right)du} }}{{\partial {x_1}}} - {A_o}\left( \phi  \right)}}{{\left( {{X_2} - {X_1} + 1} \right) - \left( {{x_2} - {x_1} + 1} \right)}}\\
& = \frac{{ - {f_k}\left( {{x_2}} \right)\frac{{\partial {x_2}}}{{\partial {x_1}}} + {f_k}\left( {{x_1}} \right)\frac{{\partial {x_1}}}{{\partial {x_1}}} - {A_o}\left( \phi  \right)}}{{\left( {{X_2} - {X_1} + 1} \right) - \left( {{x_2} - {x_1} + 1} \right)}}\\
& = \frac{{{f_k}\left( {{x_1}} \right) - {A_o}\left( \phi  \right)}}{{\left( {{X_2} - {X_1} + 1} \right) - \left( {{x_2} - {x_1} + 1} \right)}};
\end{align}
\begin{align}
\frac{{\partial {A_i}\left( \phi  \right)}}{{\partial {x_1}}} & = \frac{{\partial \frac{{\int\limits_{{x_1}}^{{x_2}} {{f_k}\left( u \right)du} }}{{\left( {{x_2} - {x_1} + 1} \right)}}}}{{\partial {x_1}}}\\
& = \frac{{\frac{{\partial \int\limits_{{x_1}}^{{x_2}} {{f_k}\left( u \right)du} }}{{\partial {x_1}}} - {A_i}\left( \phi  \right)\frac{{\partial \left( {{x_2} - {x_1} + 1} \right)}}{{\partial {x_1}}}}}{{\left( {{x_2} - {x_1} + 1} \right)}}\\
& = \frac{{\frac{{\partial \int\limits_{{x_1}}^{{x_2}} {{f_k}\left( u \right)du} }}{{\partial {x_1}}} + {A_i}\left( \phi  \right)}}{{\left( {{x_2} - {x_1} + 1} \right)}}\\
& = \frac{{{f_k}\left( {{x_2}} \right)\frac{{\partial {x_2}}}{{\partial {x_1}}} - {f_k}\left( {{x_1}} \right)\frac{{\partial {x_1}}}{{\partial {x_1}}} + {A_i}\left( \phi  \right)}}{{\left( {{x_2} - {x_1} + 1} \right)}}\\
& = \frac{{ - {f_k}\left( {{x_1}} \right) + {A_i}\left( \phi  \right)}}{{\left( {{x_2} - {x_1} + 1} \right)}};
\end{align}
\begin{align}
\frac{{\partial {{\cal L}_{\rm{OIC}}}\left( \phi  \right)}}{{\partial {x_1}}} & = \frac{{\partial {A_o}\left( \phi  \right)}}{{\partial {x_1}}} - \frac{{\partial {A_i}\left( \phi  \right)}}{{\partial {x_1}}}\\
& = \frac{{{f_k}\left( {{x_1}} \right) - {A_o}\left( \phi  \right)}}{{\left( {{X_2} - {X_1} + 1} \right) - \left( {{x_2} - {x_1} + 1} \right)}} - \frac{{{A_i}\left( \phi  \right) - {f_k}\left( {{x_1}} \right)}}{{\left( {{x_2} - {x_1} + 1} \right)}}.
\end{align}

Likewise, the gradient corresponding to the predicted segment $\phi$ w.r.t the right inner boundary $x_2$ of $\phi$ can be computed as follows:
\begin{align}
\frac{{\partial {{\cal L}_{\rm{OIC}}}\left( \phi  \right)}}{{\partial {x_2}}} = \underbrace {\frac{{{A_o}\left( \phi  \right) - {f_k}\left( {{x_2}} \right)}}{{\left( {{X_2} - {X_1} + 1} \right) - \left( {{x_2} - {x_1} + 1} \right)}}}_{\frac{{\partial {A_o}\left( \phi  \right)}}{{\partial {x_2}}}} - \underbrace {\frac{{{f_k}\left( {{x_2}} \right) - {A_i}\left( \phi  \right)}}{{\left( {{x_2} - {x_1} + 1} \right)}}}_{\frac{{\partial {A_i}\left( \phi  \right)}}{{\partial {x_2}}}}
.
\end{align}

\subsection{The Connection of the LoG Filter and Our OIC Loss} \label{sec:6}

In the paper Sec. 5.4, we have discussed how our OIC loss relates with the \textbf{Laplacian of Gaussian (LoG)} filter for blob detection \cite{lindeberg1998feature}.
When we convolve the LoG filter over a signal, the LoG filter achieves the response of maximum magnitude at the center of the target blob, provided the scale of the Laplacian is matched with the scale of the blob.

The operation of computing the OIC loss is effectively convolving the CAS with a step function as shown in the paper Fig. 5.
This step function implicated in our OIC loss can be regarded as a variant of the LoG filter for the sake of easing the network training.
The optimum (minimum OIC loss) is achieved at center of the segment whose activations in the inner area are relatively high and activations in the outer area are relatively low.
Despite the simplicity of the OIC loss, it turns out to be quite effective in practice for localizing likely action segments.

In addition, the integral of the LoG filter on the range $(-\rm{Inf},+\rm{Inf})$ is 0.
The step function implicated in our OIC loss also exhibits such characteristic.
Recall that the OIC loss is defined as the average activation in the outer area minus the average activation in the inner area.
Therefore, the value of this step function in the outer area is $\frac{1}{{{\rm{length\ of\ the\ outer\ area}}}}$ and the value of this step function in the inner area is $\frac{-1}{{{\rm{length\ of\ the\ inner\ area}}}}$.
The value of this step function in other area is 0.
Consequently, the integral on the outer area is 1 and the integral on the inner area is $-1$, and the total integral of this step function on the range $(-\rm{Inf},+\rm{Inf})$ is 0 as well.

\subsection{Details of Boundary Transformation} \label{sec:2}

\subsubsection{Details of Clipping} \label{sec:2.1}

\begin{figure}
	\centering
	\includegraphics[width=0.8\textwidth]{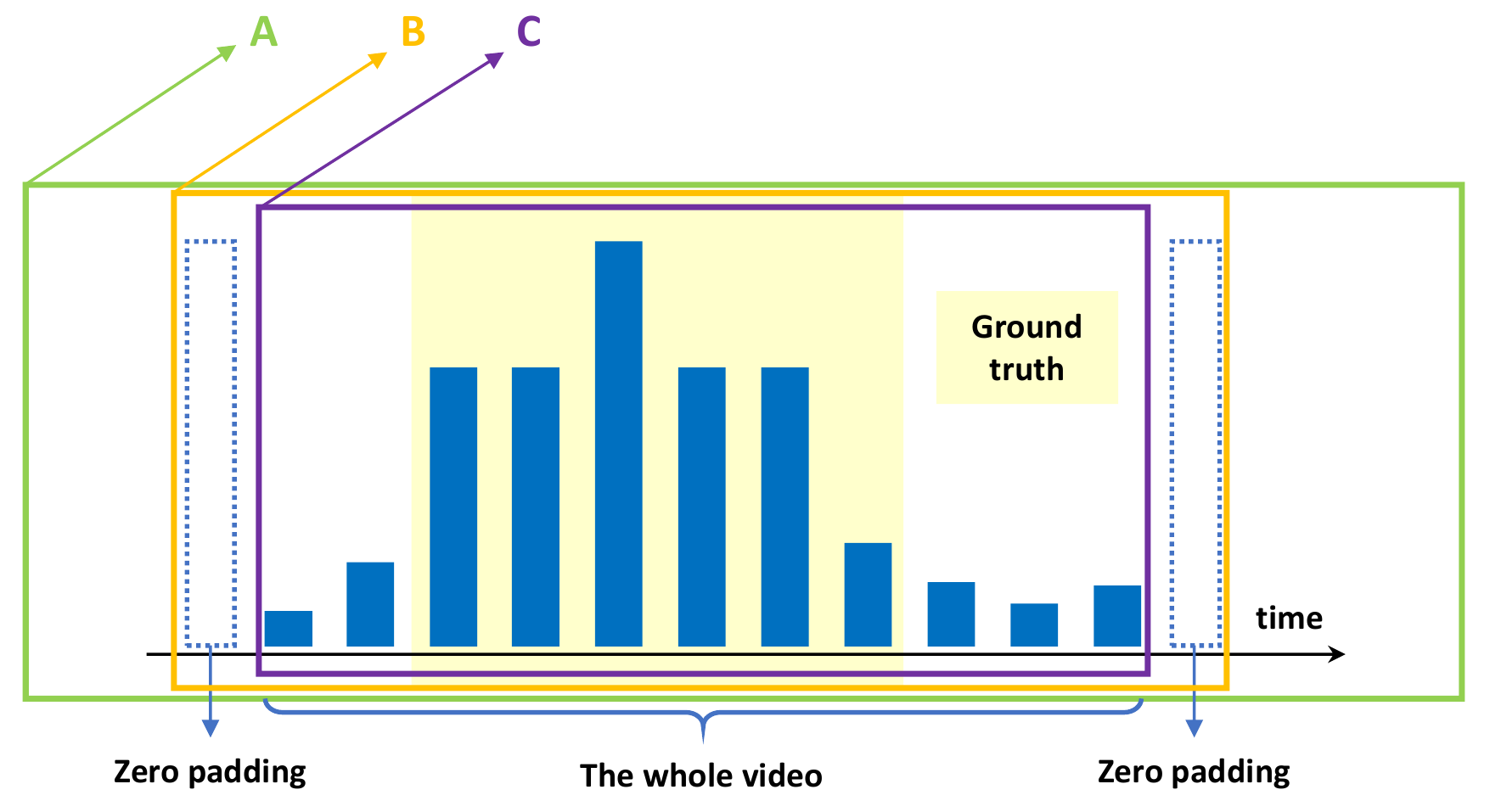}
	\caption{Visualization example of two different clipping methods. A is the predicted boundary; B is the result of clipping with zero padding (ours); C is the result of clipping without zero padding}
	\label{fig:clipping}
\end{figure}

As mentioned in the paper Sec. 4.3, when the predicted boundary $x_1$ and $x_2$ exceed the boundary of the whole video, we clip $x_1$ and $x_2$ to fit into the range of the whole video.
As an example shown in Fig.~\ref{fig:clipping}, A is the predicted inner boundary.

C is obtained by clipping A to fit into [1, $T$] ($T$ is the total number of snippets). According to Equation 2 in the paper, using this clipping method, the gradient w.r.t the inner left boundary is
\begin{align}
\frac{{\partial {{\cal L}_{\rm{OIC}}}\left( \phi  \right)}}{{\partial {x_1}}} & = \frac{{{f_k}\left( {{x_1}} \right) - {A_o}\left( \phi  \right)}}{{\left( {{X_2} - {X_1} + 1} \right) - \left( {{x_2} - {x_1} + 1} \right)}} - \frac{{{A_i}\left( \phi  \right) - {f_k}\left( {{x_1}} \right)}}{{\left( {{x_2} - {x_1} + 1} \right)}}\\
& = \frac{{{f_k}\left( {{x_1}} \right)}}{{\left( {{X_2} - {X_1} + 1} \right) - \left( {{x_2} - {x_1} + 1} \right)}} - \frac{{{A_i}\left( \phi  \right) - {f_k}\left( {{x_1}} \right)}}{{\left( {{x_2} - {x_1} + 1} \right)}}.
\end{align}
Here $\frac{{\partial {{\cal L}_{\rm{OIC}}}\left( \phi  \right)}}{{\partial {x_1}}}$ has the possibility to be positive and thus results into moving $x_1$ left.
However, we would actually like to move $x_1$ right through optimization.
This is due to that in our implementation, a segment of boundary $\left[ {{x_1},{x_2}} \right]$ includes the snippets at $x_1$ and $x_2$, and the activations at the first and the last snippets of the whole video may not be zero.

Therefore, we conduct clipping with zero-padding to obtain B: padding one snippet of zero activation at both ends of the whole video and then clipping A to fit into [0, $T+1$].
Using this clipping method, the gradient w.r.t the inner left boundary is
\begin{align}
\frac{{\partial {{\cal L}_{\rm{OIC}}}\left( \phi  \right)}}{{\partial {x_1}}} & = \frac{{{f_k}\left( {{x_1}} \right) - {A_o}\left( \phi  \right)}}{{\left( {{X_2} - {X_1} + 1} \right) - \left( {{x_2} - {x_1} + 1} \right)}} - \frac{{{A_i}\left( \phi  \right) - {f_k}\left( {{x_1}} \right)}}{{\left( {{x_2} - {x_1} + 1} \right)}}\\
& =  - \frac{{{A_i}\left( \phi  \right)}}{{\left( {{x_2} - {x_1} + 1} \right)}},
\end{align}
which is always non-positive.
Therefore, even though sometimes the predicted boundary might be too large and exceed the range of the whole video, our clipping method allows the boundary prediction model to move back to explore the potential boundary locations inside the video.

\begin{figure}
	\centering
	\includegraphics[width=0.6\textwidth]{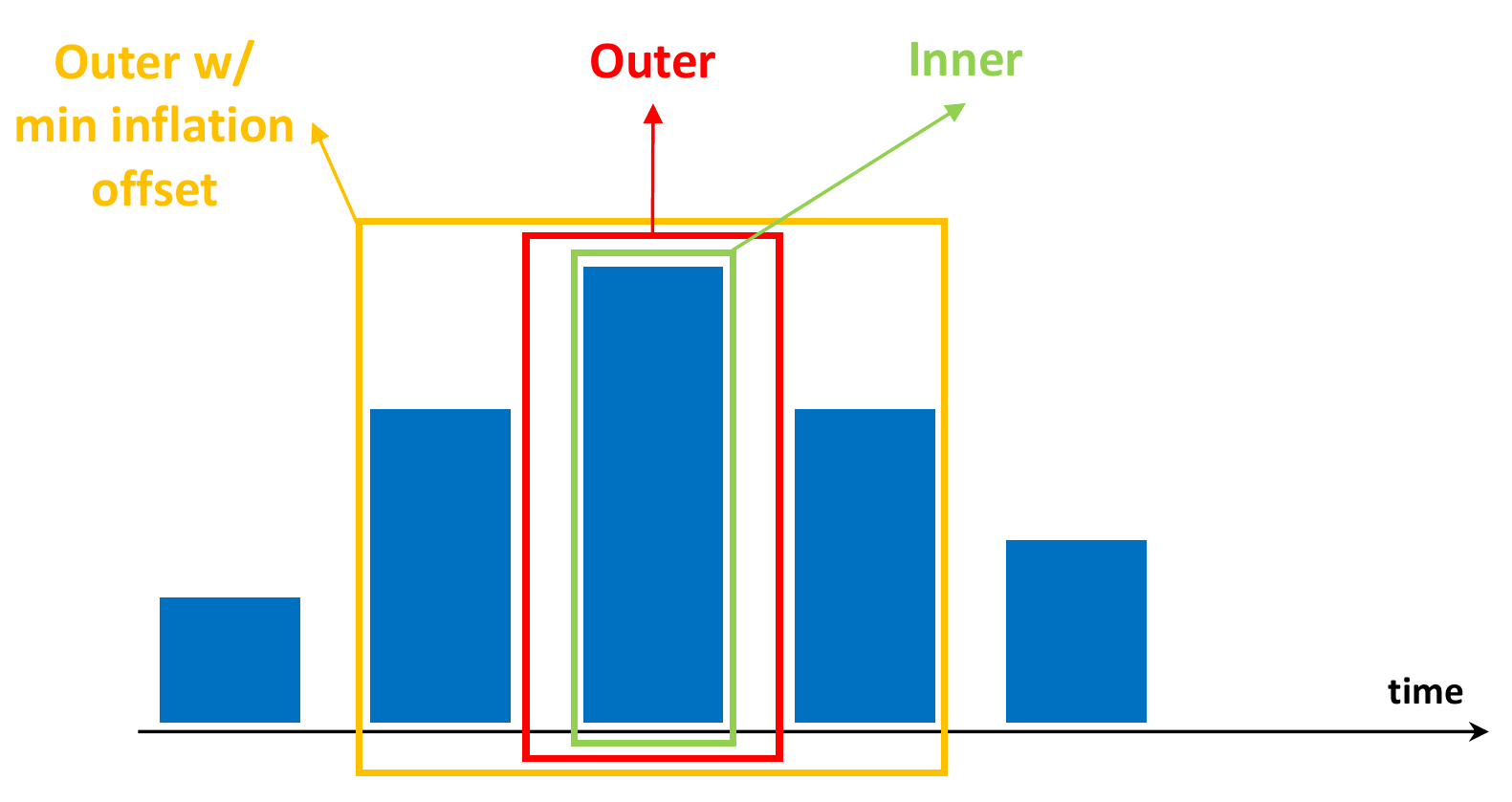}
	\caption{Visualization example of inflating the inner boundary with a certain minimum offset to obtain the outer boundary}
	\label{fig:inflation}
\end{figure}

\subsubsection{Details of Inflation} \label{sec:2.2}

As presented in the paper Sec. 4.3, in order to take into account the contextual information, we need to inflate the inner boundary $x_1$ and $x_2$ to obtain the outer boundary $X_1$ and $X_2$.
However, when the predicted segment is very short, $X_1$ would be very close to $x_1$ and $X_2$ would be very close to $x_2$.
As mentioned in the paper Sec. 3.1, when fetching the corresponding activation on CAS, we round each boundary to its nearest integer (i.e. the location of the nearest snippet).
If the inner boundary and the outer boundary are very close, they might be rounded to the same location on CAS.
Consequently, there is no outer area.
In order to keep a minimum outer area on CAS so that the model can look into the contextual information, we force $X_1$ to be not larger than $x_1 - 1$ and force $X_2$ to be not smaller than $x_2 + 1$.

As shown by an example in Fig.~\ref{fig:inflation}, the inner boundary and the outer boundary are very close.
During the OIC layer, these boundaries are rounded to be the same.
In order to look into the contrast between the inner area and the outer area, we make sure the outer boundary is extended by at least a certain minimum offset (i.e. 1 snippet).


\subsection{Details of the OIC Layer} \label{sec:3}

Alg. 1 summarizes the steps of determining whether keeping each segment prediction or not in the OIC layer during training.

\begin{algorithm}
	\textbf{Input}: For each video, one input is the CAS of shape ($K$, $T$).
	We denote the class activation at the $x$-th snippet at the CAS of the action $k$ as ${f_k}\left( x \right)$.
	For each video, another input is a set of candidate segment predictions $\Phi$ of shape ($K$, $T$, $M$). $K$ is the number of actions; $T$ is the number of snippets; $M$ is the number of anchors.
	Each prediction consists of the inner boundary and the outer boundary and the action class.
	
	\textbf{Output}: A binary-value matrix $\Omega$ of shape ($K$, $T$, $M$) to indicate whether keeping each segment prediction or not. 0 indicates removing and 1 indicates keeping.
	
	01. Initialize $\Omega$ as all zeros
	
	02. \textbf{for} each $k$ in the set of ground truth actions \textbf{do} 
	
	03. \quad \textbf{for} each temporal position $t = 1, \ldots ,T$ on CAS \textbf{do} 
	
	04. \quad \quad \textbf{if} ${f_k}\left( t \right) \ge 0.1$ \textbf{then}
	
	05. \quad \quad \quad \textbf{for} each anchor $a = 1, \ldots ,M$ \textbf{do} 
	
	06. \quad \quad \quad \quad Compute the OIC loss according to Equation 1 in the paper
	
	07. \quad \quad \quad \textbf{end for}
	
	08. \quad \quad \quad Determine the anchor $\hat{m}$ of the lowest OIC loss
	
	09. \quad \quad \quad \textbf{if} the OIC loss of $\hat{m}$ $\le$ -0.3 \textbf{then}
	
	10. \quad \quad \quad \quad $\Omega \left( {k,t,\widehat m} \right) = 1$
	
	11. \quad \quad \quad \textbf{end if}
	
	12. \quad \quad \textbf{end if}
	
	13. \quad \textbf{end for}
	
	14. \quad Update $\Omega \left( {k,:,:} \right)$ via NMS to remove highly overlapped segments according to their OIC losses
	
	15. \textbf{end for}
	
	16. \textbf{return} $\Omega$
	
	\caption{Determine whether keeping each segment prediction or not in the OIC layer during training}
	\label{alg:oic}
\end{algorithm}

\subsection{Experiments} \label{sec:4}

\subsubsection{Exploration Study}  \label{sec:4.1}

In order to determine a few hyper-parameters in AutoLoc, we first make hypothesis about the reasonable range and then explore quantitatively via grid search on the training data.
For example, $\alpha$ is used to inflate the inner boundary to obtain the outer boundary.
$\alpha$ should not be too large (larger than 1/2) which may include too many irrelevant area and $\alpha$ also should not be too small (smaller than 1/8) for the sake of taking into account sufficient contextual area preceding and succeeding the predicted action boundary (i.e. inner boundary).
Therefore, we vary $\alpha$ within a reasonable range from 1/2 to 1/8 on the training videos on THUMOS'14 and ActivityNet respectively.
As shown in Table~\ref{table:inflation-th} and Table~\ref{table:inflation-an}, within this range, the results are all acceptable and comparable on both datasets.
$\alpha$ = 1/4 is slightly better and thus we set $\alpha$ to 1/4 for all the experiments in the paper.

\begin{table}
	\begin{center}
		\caption{Temporal localization mAP (\%) when we vary the the inflation ratio $\alpha$. We train AutoLoc on the THUMOS'14 val set and evaluate on the THUMOS'14 val set.}
		\label{table:inflation-th}
		\begin{tabular}{c|ccccc}
			IoU threshold & 0.3           & 0.4           & 0.5           & 0.6           & 0.7          \\ \hline
			$\alpha$ = 1/2           & 39.6          & 33.1          & 25.8          & 16.5          & 8.3          \\
			$\alpha$ = 1/4          & \textbf{41.2} & \textbf{33.8} & \textbf{25.8} & \textbf{17.9} & \textbf{9.7} \\
			$\alpha$ = 1/8         & 40.4          & 32.8          & 24.6          & 16.4          & 7.8         
		\end{tabular}
	\end{center}
\end{table}

\begin{table}
	\begin{center}
		\caption{Temporal localization mAP (\%) when we vary the the inflation ratio $\alpha$. We train AutoLoc on the ActivityNet train set and evaluate on the ActivityNet train set.}
		\label{table:inflation-an}
		\begin{tabular}{c|cccccccccc|c}
			IoU threshold & 0.5           & 0.55          & 0.6           & 0.65          & 0.7           & 0.75          & 0.8           & 0.85          & 0.9          & 0.95         & Avg           \\ \hline
			$\alpha$ = 1/2           & 31.2          & 27.5          & \textbf{24.7} & 21.7          & 18.8          & 15.7          & 12.7          & 9.7           & 6.4          & 2.6          & 17.1          \\
			$\alpha$ = 1/4          & \textbf{31.6} & 27.8          & 24.6          & \textbf{22.0} & \textbf{19.2} & \textbf{16.4} & \textbf{13.3} & \textbf{10.3} & \textbf{6.6} & 2.5          & \textbf{17.4} \\
			$\alpha$ = 1/8         & 31.5          & \textbf{27.9} & 24.4          & 21.5          & 18.9          & 15.9          & 13.0          & 10.1          & 6.5          & \textbf{2.8} & 17.2         
		\end{tabular}
	\end{center}
\end{table}

\subsubsection{Formulation of the Inner Only Loss} \label{sec:4.2}

In the paper Sec. 5.4, we have investigated changing the OIC loss in AutoLoc to \textbf{Inner Only Loss}, which only encourages the high activations inside the segment while does not look into the contextual area.
The \textbf{Inner Only Loss} is defined as:
\begin{align}
{\cal L}_{\rm{inner}}\left( \phi  \right) =  - {A_i}\left( \phi  \right) =  - \frac{{\int\limits_{{x_1}}^{{x_2}} {{f_k}\left( u \right)du} }}{{\left( {{x_2} - {x_1} + 1} \right)}}.
\end{align}

Since the \textbf{Inner Only Loss} only considers the inner boundary, we only need to back-propagate the gradients w.r.t the inner boundary, which can computed as follows:
\begin{align}
\frac{{\partial {{\cal L}_{\rm{inner}}}\left( \phi  \right)}}{{\partial {x_1}}} =  - \frac{{{A_i}\left( \phi  \right) - {f_k}\left( {{x_1}} \right)}}{{\left( {{x_2} - {x_1} + 1} \right)}};\\
\frac{{\partial {\cal L}_{\rm{inner}}\left( \phi  \right)}}{{\partial {x_2}}} =  - \frac{{{f_k}\left( {{x_2}} \right) - {A_i}\left( \phi  \right)}}{{\left( {{x_2} - {x_1} + 1} \right)}}.
\end{align}

\subsubsection{The Advantage of Combining Anchor Mechanism and Boundary Regression Compared to Enumeration with OIC Selection} \label{sec:4.3}

\begin{figure}
	\centering
	\includegraphics[width=.6\textwidth]{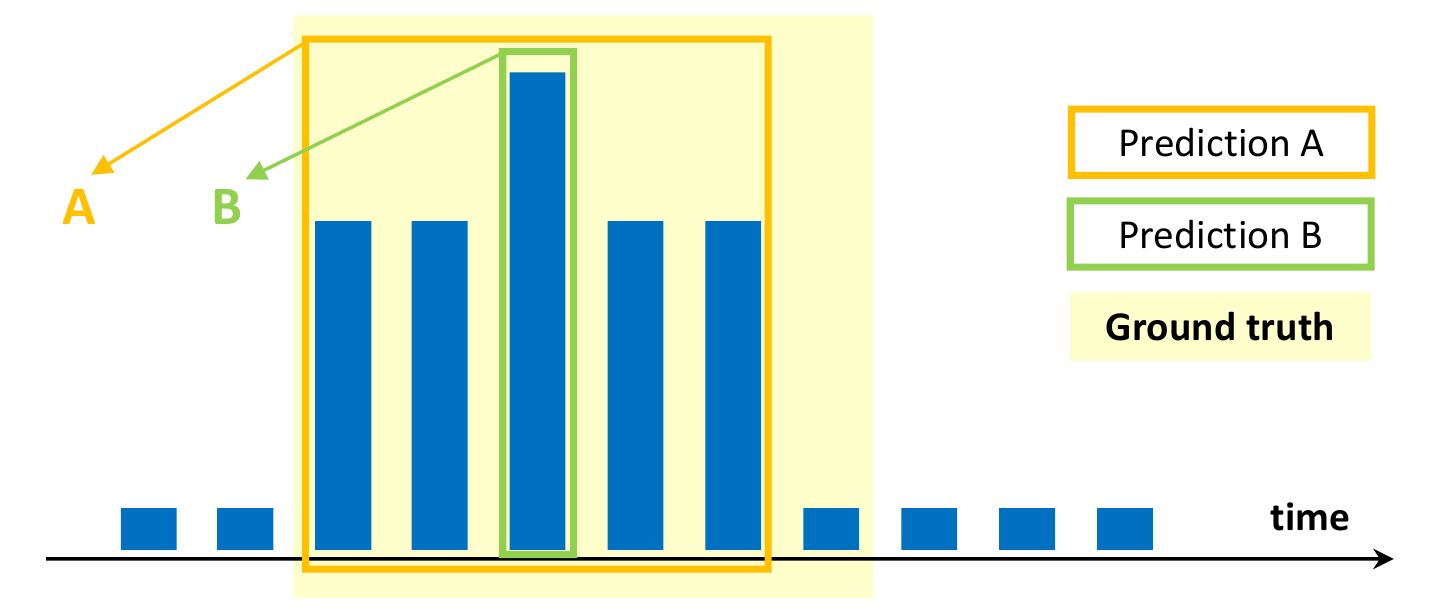}
	\caption{A visualization example to illustrate the advantage of combining anchor mechanism and boundary regression compared to enumeration with OIC selection}
	\label{fig:enum}
\end{figure}

As shown in Table 3, even without training on the training videos, \textbf{Direct Optimization} outperforms \textbf{OIC Selection}.
This implies the advantage of combining anchor mechanism and boundary regression over performing enumeration with the OIC selection.

\textbf{Direct Optimization} and \textbf{OIC Selection} both generate a set of candidate segments first and then leverage the OIC loss to obtain the final predictions.
\textbf{OIC Selection} enumerates a large set of possible segments and uses the OIC loss to select the likely ones while \textbf{OIC Selection} sometimes keeps quite a few false alarms. 
But \textbf{Direct Optimization} generates a small set of anchors at each position on CAS and then select the one with the lowest OIC loss, which has the most likely temporal scale, and then \textbf{Direct Optimization} refines the boundary of the selected anchor through optimization.
Consequently, \textbf{Direct Optimization} compares segments cross various scales and thus is less likely to fall into local minimum.
The precise boundary can be found through the optimization afterwards.
We give a visualization example in Fig.~\ref{fig:enum} to help illustration.

As for \textbf{OIC Selection}, it might enumerate both prediction A and B.
A is better aligned to the ground truth boundary.
B localizes a peak because the activation at the peak is indeed much higher than the activations in its contextual area.
But B might be a solution at the local minimum, compared to A which has even lower OIC loss.
However, since A and B are not highly overlapped (their overlap IoU is 0.2), both A and B are likely to be kept by \textbf{OIC Selection} even after NMS.
As for \textbf{Direct Optimization}, in the case that the model has two anchors (one is regressed as A and another one is regressed as B), the model keeps B and removes A directly.


In addition, considering the speed, enumeration is not practical for long videos.
But for the TAL task, videos are usually untrimmed and long.

\clearpage

\bibliographystyle{splncs04}
\bibliography{egbib}

\end{document}